%% file: 00-main.tex
\documentclass[sigconf]{acmart}

\usepackage{multicol}
\usepackage{multirow}

\usepackage{algorithm}
\usepackage{algpseudocode}
\algnewcommand{\LeftComment}[1]{\State \(\triangleright\) #1}

\algtext*{EndWhile}
\algtext*{EndIf}
\algtext*{EndFor}
\algtext*{EndFunction}

\newcommand{\eg}{{\textit{e.g.}}}
\newcommand{\ie}{{\textit{i.e.}}}
\newcommand{\etal}{{\textit{et al.}}}

\usepackage{array}
\newcommand{\PreserveBackslash}[1]{\let\temp=\\#1\let\\=\temp}
\newcolumntype{C}[1]{>{\PreserveBackslash\centering}p{#1}}

\AtBeginDocument{%
  \providecommand\BibTeX{{%
    \normalfont B\kern-0.5em{\scshape i\kern-0.25em b}\kern-0.8em\TeX}}}

\setcopyright{none}

\renewcommand\footnotetextcopyrightpermission[1]{} 
\begin{document}

\title{ONE-NAS: An Online NeuroEvolution based Neural Architecture Search for Time Series Forecasting}

\author{Zimeng Lyu}
\email{zimenglyu@mail.rit.edu}

\affiliation{%
  \institution{Rochester Institute of Technology}
  \city{Rochester}
  \state{New York}
  \country{USA}}
  
\author{Travis Desell}
\email{tjdvse@rit.edu}

\affiliation{%
  \institution{Rochester Institute of Technology}
  \city{Rochester}
  \state{New York}
  \country{USA}}

\renewcommand{\shortauthors}{Lyu, et al.}


\input{01-abstract}

\maketitle
\pagestyle{plain} 
\input{02-introduction}
\input{03-relatedwork}
\input{04-method}

\input{05-results}
\input{06-conclusion}


\bibliographystyle{ACM-Reference-Format}
\bibliography{99-reference}


\end{document}

%% file: 01-abstract.tex
\begin{abstract}

Time series forecasting (TSF) is one of the most important tasks in data science, as accurate time series (TS) predictions can drive and advance a wide variety of domains including finance, transportation, health care, and power systems. However, real-world utilization of machine learning (ML) models for TSF suffers due to pretrained models being able to learn and adapt to unpredictable patterns as previously unseen data arrives over longer time scales. To address this, models must be periodically retained or redesigned, which takes significant human and computational resources. This work presents the Online NeuroEvolution based Neural Architecture Search (ONE-NAS) algorithm, which to the authors' knowledge is the first neural architecture search algorithm capable of automatically designing and training new recurrent neural networks (RNNs) in an online setting. Without any pretraining, ONE-NAS utilizes populations of RNNs which are continuously updated with new network structures and weights in response to new multivariate input data. ONE-NAS is tested on real-world large-scale multivariate wind turbine data as well a univariate Dow Jones Industrial Average (DJIA) dataset, and is shown to outperform traditional statistical time series forecasting, including naive, moving average, and exponential smoothing methods, as well as state of the art online ARIMA strategies.  




\end{abstract}

\ccsdesc[500]{Theory of computation~Online algorithms}
\ccsdesc[500]{Computing methodologies~Neural networks}
\ccsdesc[500]{Applied computing~Forecasting}

\keywords{NeuroEvolution, Online Algorithms, Time Series Forecasting, Recurrent Neural Networks, Neural Architecture Search}

%% file: 02-introduction.tex
\section{Introduction}

Time series forecasting (TSF) is one of the most important tasks in data science, as accurate time series (TS) predictions can drive and advance a wide variety of domains including finance~\cite{cao2019financial, vlasenko2019novel}, transportation~\cite{ghosh2005time, wu2012online}, health care~\cite{bhatnagar2012forecasting, zinouri2018modelling}, and power systems~\cite{osegi2020using,al2020electrical}. A major problem in TSF is that models often do not perform well over long time periods due to changing distributions between their original training data and new inference data, as the properties or important features of the data change over time~\cite{guo2016robust}. Due to this, these models can only be used for predictions for a reasonable period of time into the future, and then need to be retrained or redesigned as more recent data arrives so that the model can learn previously unseen patterns. This repeated process is time consuming especially as the architectures of artificial neural networks (ANNs) become more and more complicated to handle multi-variate large-scale time series data.

While there has been significant research into TSF where models are first trained offline~\cite{mahalakshmi2016survey,lim2021time,torres2021deep}, there is significantly less research into online TSF, where predictions must be made in real time and models are pretrained on prior data, making the problem significantly more challenging. Recent strategies for online TSF include variations on online ARIMA~\cite{liu2016online,anava2013online} for univariate time series forecasting, or for online time series forecasting with machine learning~\cite{guo2016robust}. Online ARIMA and ARMA models are mostly limited to univariate datasets, which have limitations to most real-world TSF applications which are multi-variate. For multivariate TSF, recurrent neural networks (RNNs) can be used, however finding the right depth or capacity of the RNN is challenging, requiring significant effort to avoid vanishing gradients (when models are too complex) and not being able to learn complex patterns (when models are too simple)~\cite{sahoo2017online}. Additionally, online learning algorithms also need to deal with the problem of catastrophic forgetting~\cite{mccloskey1989catastrophic, abraham2005memory}, where important historical information can be forgotten over time\footnote{Offline strategies suffer the reverse problem, as described above. They fail to learn from new data, and eventually the distributions and features that the models were trained on do not represent the data they are performing inference on.}.

Online time series forecasting strategies which incorporate neuroevolution, on the other hand, can tackle those challenges. Many neuroevolution algorithms, such as those related to the popular NeuroEvolution of Augmenting Topologies (NEAT) algorithm~\cite{stanley2002evolving}, start with seed genomes with a minimal structure. During the evolutionary process, networks grow to gradually adapt to the complexity of the dataset, which reduces the effort required to hand tune and design the networks. Architectures can be updated and trained as new data arrives, while populations can retain older networks which can contain valuable historical information to reduce or prevent catastrophic forgetting.

Therefore, we propose a novel online NeuroEvolution (NE) Neural Architecture Search (NAS) method that evolves Recurrent Neural Networks (RNNs) for time series data prediction (ONE-NAS). To authors' knowledge this is the first algorithm capable of the online evolution of RNNs for time series data predictions. It provides numerous benefits over traditional fixed RNN architectures and even other modern NE strategies, allowing the RNN architectures to be continually updated in response to new input data, as well as gracefully handling the effect of unpredictable events. It is able to more quickly be trained in response to incoming data streams, as opposed to alternative methods that require significant offline time training using previously gathered training data sets. As historical time series data can have important temporal information, we train the child genomes with randomly selected historical data, so different historical temporal information is saved in different children. Eventually the genomes with useful weights and structures survive and others gradually die off.

Our results evaluate ONE-NAS against traditional statistical online TSF methods, on highly challenging, noisy, real world multivariate time series data from wind turbine sensors, as well as against modern online ARMIA based methods on univariate Dow Jones Industrial Average (DJIA) data, and show significant improvements in accuracy over these methods. Further, ONE-NAS utilizes a distributed, scalable algorithm and is shown to be able to operate efficiently in real time over short time scales.  We also show that by utilizing multiple islands which are periodically repopulated to prevent stagnation in ONE-NAS, we significantly improve performance of the algorithm.

%% file: 03-relatedwork.tex
\section{Related Work}

Most online time series forecasting models use fixed mathematical models or fixed NN topologies. These models adapt to different datasets by training and updating the weights as new data arrives. For statistical approaches, variations on online Autoregressive Integrated Moving Average (ARIMA) models have been proposed for online time series forecasting by Liu \etal ~\cite{liu2016online}, as well as for anomaly detection by Kozitsin \etal~\cite{kozitsin2021online}. Online ARIMA has also been used for unsupervised anomaly detection~\cite{schmidt2018unsupervised,schmidt2018unsupervised}.
The Autoregressive moving average (ARMA) model and Seasonal AutoRegressive Integrated Moving Average (SARIMA) proposed by Anava~\etal~and~He~\etal~have also been used for time series forecasting~\cite{anava2013online,han2018multivariate}, with 
Anava~\etal~ additionally proposing an autoregressive (AR) model for time series forecasting with missing data~\cite{anava2015online}.

For neural network based approaches, Guo~\etal~proposed an adaptive gradient learning method for training recurrent neural networks (RNNs) for time series forecasting in the presence of anomalies and change points~\cite{guo2016robust}. Their proposed model dynamically slides over the input data and updates the weights of an RNN network to reduce the weights of suspicious anomalies. Yang~\etal~use RoAdam (Robust Adam) to train long short-term memory (LSTM) RNNs for online time series prediction in the presence of outliers~\cite{yang2017robust}. To reduces the adverse effect of outliers on the LSTM networks, they use an adaptive learning rate which reduces when the relative prediction error is reduced, and is increased when the relative error improves~\cite{yang2017robust}. Wang~\etal~design an online sequential extreme learning machine with kernels (OS-ELMK) for nonstationary time series forecasting. A fixed memory online time series prediction scheme is used to adapt the weights in response to incoming data~\cite{wang2014online}.

Neuroevolution has been widely used for time series prediction and neural architecture search in offline scenarios~\cite{ororbia2018using, lyu2021Improving, lyu2021experimental}. However, online neuroevolution has only been rarely investigated, with a few algorithms designed for games or simulators that involves real-time interactions, such as an online car racing simulator~\cite{cardamone2010learning}, online video games~\cite{agogino2000online,stanley2005real}, and robot controllers~\cite{galassi2016evolutionary}. These online NE NAS algorithms are based on the venerable NeuroEvolution of Augmenting Topologies (NEAT) algorithm~\cite{stanley2002evolving} and start with minimal networks and evolve topology and weights during the evolutionary process.
Agogino~\etal~developed online NE algorithm that evolve feed-forward NNs to play against humans in real-time~\cite{agogino2000online}. After a short period of training and evaluation offline, the NN agents continually evolve online and potentially improve performance significantly, even adapting to novel situations brought about by changing strategies in the opponent and the game layout. Stanley~\etal~designed an real-time online NE video game, Neuroevolving Robotic Operatives (NERO), based on real-time Neuroevolution of Augmenting Topologies (rtNEAT) method~\cite{stanley2005real}. Players train an NE agent team real-time to play with the team trained by another player~\cite{stanley2005real}. Cardamone~\etal~developed online car racing simulator based on NEAT~\cite{stanley2002evolving} and rtNEAT \cite{stanley2005real}, combined with four evaluation strategies ($\epsilon$-greedy~\cite{sutton2018reinforcement},
$\epsilon$-greedy-improved, softmax, and interval-based). Their algorithm evolves car drivers from scratch, which can be transferred to other race car tracks. The online NE simulator was also shown to be able to outperform offline models~\cite{cardamone2010learning}.

To our knowledge, none of these online neuroevolution algorithms were capable of evolving recurrent neural networks, nor have any been developed to perform time series forecasting, making ONE-NAS the first of its kind.

%% file: 04-method.tex
\section{Methodology}
\label{sec:methodology}

The proposed work is utilizes components from the Evolutionary Exploration of Augmenting Memory Models (EXAMM) algorithm~\cite{ororbia2019examm} as its core and utilizes its mutation, crossover and training operations in online scenarios. EXAMM is a distributed NE algorithm that evolves progressively larger RNNs for large-scale, multivariate, real-world TSF~\cite{elsaid2020improving,elsaid2020neuro}.  EXAMM evolves RNN architectures consisting of varying recurrent connections and memory cells through a series of mutation and crossover (reproduction) operations. Memory cells are selected from a library of $\Delta$-RNN units~\cite{ororbia2017diff}, gated recurrent units (GRUs)~\cite{chung2014empirical}, long short-term memory cells (LSTMs)~\cite{hochreiter1997long}, minimal gated units (MGUs)~\cite{zhou2016minimal}, and update-gate RNN cells (UGRNNs)~\cite{collins2016capacity}. ONE-NAS utilizes EXAMM's parallel asynchronous strategy which naturally load balances itself and allows for decoupling population size from the number of workers during each generation~\cite{ororbia2019examm}. Generated offspring inherit their weights from their parents, which can significantly reduce the time needed for their training and evaluation~\cite{lyu2021experimental}. It has been shown that EXAMM can swiftly adapt RNNs in transfer learning scenarios, even when the input and output data streams are changed~\cite{elsaid2020improving}~\cite{elsaid2020neuro}, which served as a preliminary motivation and justification for being able to adapt and evolve RNNs for TSF in online scenarios.


\subsection{ONE-NAS}
\label{sec:one-nas}

\begin{figure*}[t]
    \centering
	\includegraphics[width=0.75\textwidth]{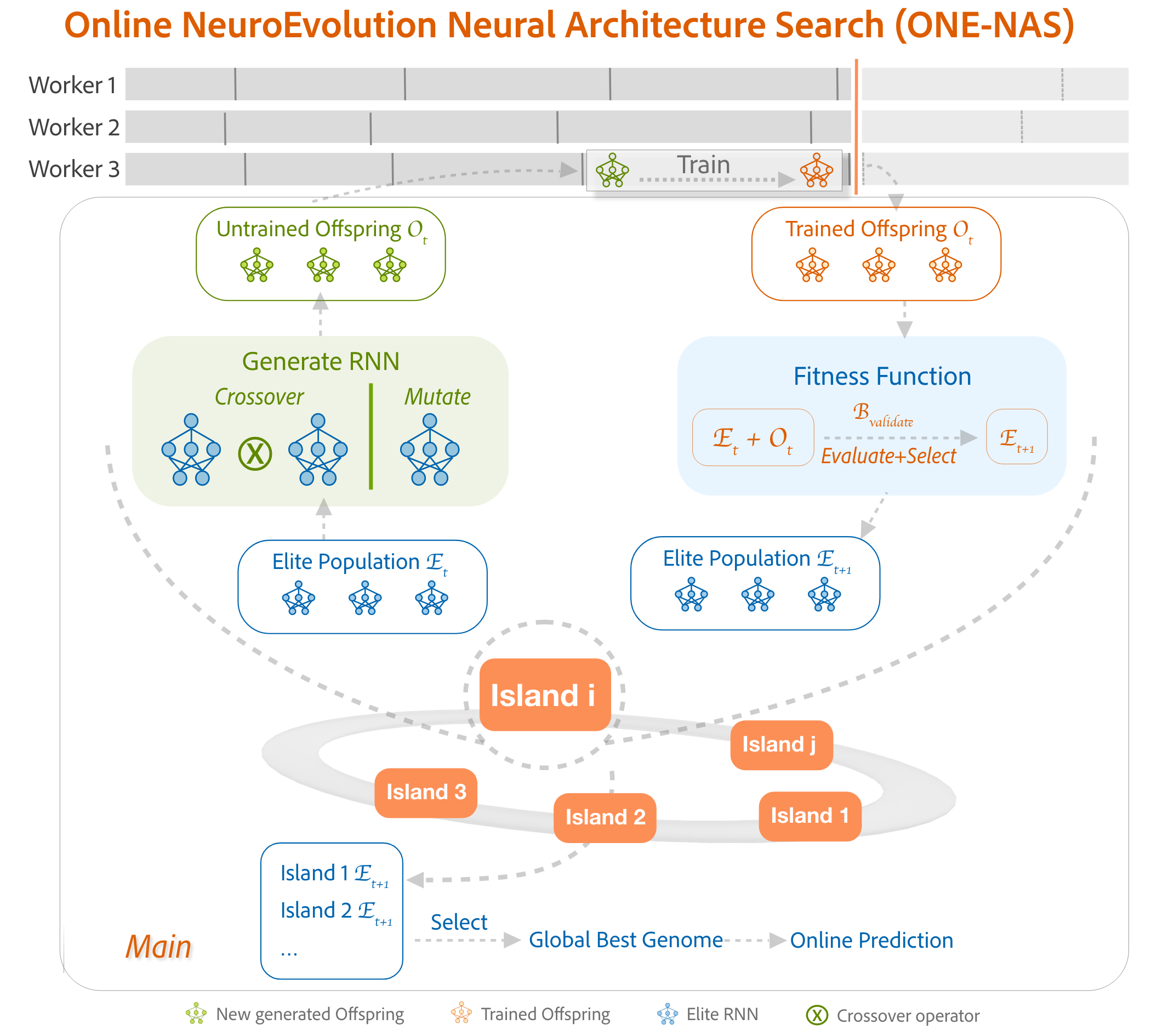}
	\caption{This figure represents one generation of the ONE-NAS online neuroevolution process. The global best genome performs predictions concurrently with distributed genome generation and evaluation. }
	\label{fig:onenas}
\end{figure*}

Figure~\ref{fig:onenas}~presents a high level view of the asynchronous, distributed, online ONE-NAS algorithm, and Algorithms~\ref{alg:main},~\ref{alg:data}, and ~\ref{alg:repopulation} present pseudocode for the algorithm. This paper evaluates two versions, one which utilizes a single island, and the second which utilizes multiple islands and a repopulation strategy. ONE-NAS concurrently evolves and trains new RNNs while performing online time series data prediction. ONE-NAS does not require pre-training on any data before the online NE process. 

ONE-NAS's online NE process starts with a minimal seed genome (an RNN which consists of online input nodes fully connected to output nodes, with no hidden layers), which serves as an initial genome. It progressively evolves generations of genomes by:

\begin{itemize}
    \item Selecting a set of $n$ best genomes as an elite population, defined as $E_t$ for generation $t$.
    \item Using the elite population to generate an additional $m$ genomes for the next generation through mutation and crossover from only elite parents, defined as $O_{t+1}$.
    \item Performing selection on the elite population and adding them to the next generation.
    \item Training all the non-elite genomes in the new generation for a specified number of backpropagation epochs.
    \item Evaluating all the generation's genomes on recent validation data to calculate their fitnesses.
\end{itemize}

ONE-NAS is asynchronous, with generated genomes for each generation being trained by worker processes using a work stealing strategy~\cite{mit-cilk-94} where each worker independently requests more genomes for training when it finishes training its previous requested genome. This strategy is naturally load balanced as workers do not block on any other worker for training, and scales up to a number of workers equal to the population size.

Each generation lasts for a specified period of time steps, $p$ (in this work, $p = 25$), which provides a subsequence of time series data. The best genome from the previous generation performs online predictions of the new subsequence ($B_{next}$)as it arrives, while concurrently the new generation of genomes is generated and trained. At the end of a generation, this new subsequence of data is added to ONE-NAS's historical training data.

During a generation, the generated genomes $O_t$ are trained on a randomly selected set of $B_{train}$ subsequences of historical training data, after which the entire population (including elite $E_t$) will be validated on the most recent $B_{validation}$ subsequences. Each genomes fitness, calculated as the mean squared error loss over $B_{validation}$, is then used to select the next elite population $E_{t+1}$. The best genome in $E_{t+1}$ is used for online prediction for current generation. 

Beacuse of this, while $O_t$ are trained using backpropagation on batches drawn from the historical data, the RNNs in $E_t$ do not continue to be trained. As not all RNNs in $E_t$ will preform better than those in $O_t$, ``obsolete'' RNNs will naturally die off over time, however RNNs with strong performance will remain.  Also note that one of EXAMM's mutation operations utilized by ONE-NAS is a clone operation, which allows for a duplicate of an elite parent to be retained and trained (re-using its parent's weights due to EXAMM's Lamarckian weight inheritance strategy) in the next generation.

\begin{algorithm}[t]
    \begin{algorithmic}
        \Function{ONE-NAS MAIN}{}

            \For{$t$ in $generations$}
                \LeftComment{Perform predictions concurrently in a new thread,}
                \LeftComment{which returns the next time subsequence when}
                \LeftComment{complete}
                \State $B_{next}$ = $globalBestGenome$.onlinePredict() 
                \State{}
                
                \If{islandRepopulation}
                    \LeftComment{Perform extinction and repopulation}
                    \If{t \% extinctFreq == 0}
                    \State $worstIsland$ = rankIslands().last()
                    \State repopulate($worstIsland$, $globalBestGenome$)
                    \EndIf
                \EndIf 
                \State{}
                
                \LeftComment{Generate genomes in main process}
                \State $O_t$ = MPIMain.generateGenomes($E_t$)
                \LeftComment{Process genomes asynchronously in parallel on worker}
                \LeftComment{processes}
                \State $B_{validation}$ = getValidationData($t$, $numValidationSets$)
                \For{genome $g$ in $O_t$}
                    \LeftComment{Each worker randomly selects different training data}
                    \State $B_{train}$ = getTrainingData($t$, $numTrainingSets$)
                    \State MPIWorker.trainGenomes($g$, $B_{validation}$, $B_{train}$)
                \EndFor
                \LeftComment{Main process waits for all workers to complete training}
                \State MPI-Barrier()
                \State{}
                
                \LeftComment{Evaluate genomes for next generation}
                \State $E_{t+1}$ = selectElite($E_t$, $O_t$, $B_{validation}$)
                \State $globalBestGenome$ = getGlobalBestGenome()
                \State{}
                
                \LeftComment{Wait for the online prediction thread for the generation}
                \LeftComment{to complete, which returns $B_{next}$}
                \State $wait$($B_{next}$) 
                \State{}
                
                \LeftComment{Get the latest subsequence of data and add it to the}
                \LeftComment{historical data}
                \State timeSeriesSets.append($B_{next}$)
            \EndFor
        \EndFunction
    \end{algorithmic}
    \caption{\label{alg:main} {ONE-NAS}}
\end{algorithm}

\begin{algorithm}[t]
    \begin{algorithmic}
        \Function{GetValidationData}{$t$, $numValidationSets$}
            \State $B_{train}$ = timeSeriesSets[$t-numValidationSets$ : $t$]
        \EndFunction
        
        \Function{getTrainingData}{$t$}
            \State timeSeriesSets[0 : $t-numValidationSets$].shuffle()
            \State $B_{validation}$ = timeSeriesSets[0 : $numTrainSets$]
        \EndFunction
    \end{algorithmic}
    \caption{\label{alg:data} {Data Selection Methods}}
\end{algorithm}




\subsection{ONE-NAS Island Repopulation Strategy}

As an extension to ONE-NAS, we add a version which includes islands which are periodically repopulated after extinction events. In the ONE-NAS Island Repopulation strategy, there are $m$ islands, with each island having their own set of elite $E$ and generated $O$ genomes. Each island evolves similarly as in ONE-NAS, except crossover now has two forms, \emph{intra-island crossover}, where both parents are selected from the elite population in the same island, and \emph{inter-island crossover}, where when generating a new child for an island, one parent is selected from that island's elite population, and the other is the best genome from a randomly selected different island. This strategy helps islands grow into their own niches and helps better explore the search space.  It also provides robustness to variations in the online data as it arrives.

However, it is possible for islands to get stuck in local optima and stagnate. We utilize the island extinction and repopulation presented by Lyu \etal~\cite{lyu2021Improving}, which periodically selects the worst performing island and removes all its genomes, replacing them with mutations of the global best genome from the search at that time. This strategy has been shown to even further improve the performance of island based neuroevolution strategies. For ONE-NAS's island repopulation strategy, after the repopulating island is erased, only the elite population $E_t$ of the island is filled with mutations of the global best genome, while the generated population $O_t$ is empty. At next generation $t+1$, $O_{t+1}$ is filled with child genomes generated by the repopulated $E_t$.

\begin{algorithm}[t]
    \begin{algorithmic}
        \Function{Repopulate}{$island$, $globalBestGenome$}
            \State $island$.erase()
            \For{$eliteCapacity$}
                \State $mutatedGenome$ = $globalBestGenome$.mutate()
                \State $island$.$E_t$.insert($mutatedGenome$)
            \EndFor
        \EndFunction
    \end{algorithmic}
    \caption{\label{alg:repopulation} {ONE-NAS Island Repopulation}}
\end{algorithm}

%% file: 05-results.tex
\begin{figure*}[t]
    \centering
	\includegraphics[width=0.975\textwidth]{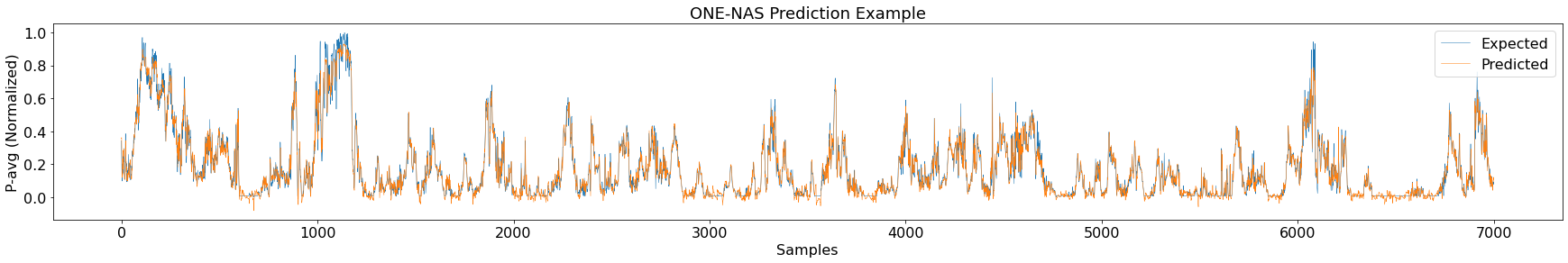}
	\caption{The \emph{average active power} parameter from the wind dataset used in this work, as well as example of ONE-NAS's predictions on this dataset.}
	\label{fig:sample}
\end{figure*}

\section{Datasets}
This work utilized two real-world data sets for predicting time series data with RNNs. The first was wind turbine engine data collected and made available by ENGIE's La Haute Borne open data windfarm\footnote{https://opendata-renewables.engie.com} which was gathered between 2017 and 2020. This wind dataset is very long, multivariate (with 22 parameters), non-seasonal, and the parameter recordings are not independent. The wind turbine data consists of readings every 10 minutes from 2013 to 2020. \emph{Average Active Power} was selected as output parameter to forecast for the wind turbine data set. Figure~\ref{fig:sample}~provides an example of the noisiness and complexity of the output parameter, as well as an example of ONE-NAS's accurate predictions on this data. The second dataset is the daily index of Dow Jones Industrial Average (DJIA) from the years 1885-1962. This dataset is univariate with 35701 samples. Both datasets contain raw and abnormal data points, and have not been cleaned, so spikes and outliers have not been removed or smoothed in the datasest.



\section{Experimental Design}

Research has shown that utilizing shorter subsequences of time series data during training can improving an RNN's convergence rate and overall performance~\cite{lyu2021neuroevolution}. For these experiments, the original datasets were divided into subsequences of $25$ timesteps each. During each ONE-NAS generation, each newly generated genome was trained on $600$ randomly selected subsequences from the historical data pool and then validated using the most recent $100$ subsequences of historical data. There is no overlap between training and validation data (the most recent $100$ subsequences are added into the historical pool after being used for validation). All the experiments for wind dataset were run for $2000$ generations, which represents one singular pass over the entire wind data time series. 

In the ONE-NAS experiments, during each generation, $50$ elite genomes from the previous generation were retained, and the elite genomes were used to generate $100$ new genomes using a mutation rate of $0.4$ and crossover rate of $0.6$. Each of the $100$ non-elite genomes in the new generation were trained in a worker process for $10$ backpropagation epochs, with the first $5$ epoch trained on the original subsequence data, and then in each of the last $5$ epochs, $10\%$ Gaussian noise was added with mean and standard deviation of the sliced data as an augmentation technique to help prevent over fitting~\cite{bishop1995training, bishop1995neural}. ONE-NAS with Island Repopulation utilized $10$, $20$, $30$ or $40$ islands, with each having its own elite population of $5$ genomes which generated an additional $10$ genomes per generation. New genomes were generated with a mutation rate of $0.3$, inter-island crossover rate of $0.4$, and intra-island crossover rate of $0.3$. 

For genome generation, $10$ out of EXAMM's $11$ mutation operations were utilized (all except for \emph{split edge}), and each was chosen with a uniform 10\% chance. ONE-NAS generated new nodes were by selecting from EXAMM's library of simple neurons, $\Delta$-RNN, GRU, LSTM, MGU, and UGRNN memory cells uniformly at random. Recurrent connections could span any time-skip generated randomly between $\mathcal{U}(1,10)$. Backpropagation (BP) through time was run with a learning rate of $\eta = 0.001$ and used Nesterov momentum with $\mu = 0.9$. For the memory cells with forget gates, the forget gate bias had a value of $1.0$ added to it (motivated by \cite{jozefowicz2015empirical}).  To prevent exploding gradients, gradient scaling~\cite{pascanu2013difficulty} was used when the norm of the gradient exceeded a threshold of $1.0$. To combat vanishing gradients, gradient boosting (the opposite of scaling) was used when the gradient norm was below $0.05$. These parameters have been selected as suggested in prior published work on EXAMM.

\section{Results}
\label{sec:results}

Each experiment was repeated $10$ times using Rochester Institute of Technology's research computing systems. This system consists of 2304 Intel® Xeon® Gold 6150 CPU 2.70GHz cores and 24 TB RAM, with compute nodes running the RedHat Enterprise Linux 7 system. Each experiment utilized 16 cores.

\subsection{Comparison to Classical TSF Methods}
To test the performance of ONE-NAS, it was first compared to classical TSF methods: naive prediction, moving average prediction, and simple exponential smoothing. These methods were selected because they are easily capable of online prediction, whereas ARMIA methods require the entire dataset \emph{a priori}. Additionally, for online TSF, research has shown simple classical methods, such as linear methods and exponential smoothing, can outperform complex and sophisticated methods~\cite{makridakis2018statistical}. Later, in Section~\ref{sec:online_arima}, we compare ONE-NAS directly to a state of the art online ARIMA method which does not require pretraining~\cite{liu2016online}.



Naive prediction simply uses the parameter's previous value, $x_{t-1}$ as the predicted value, $\hat{y_t}$, for the next time step:
\begin{equation}
    \hat{y_t} = x_{t-1}
\end{equation}

Moving average prediction uses the average of the last $n$ time steps as the predicted for the next time step, where $n$ is the moving average data smoothing window (a hyperparameter).
\begin{equation}
    \hat{y_t} = 1/n * \sum_{n=1}^{n} x_{t-n}
\end{equation}

Simple Exponential smoothing (Holt linear), computes a running average of the previously seen parameters, where $\alpha$ is the smoothing factor, and $0< \alpha <1$:
\begin{equation}
    \hat{y_t} = \alpha * x_{t-1} + (1 - \alpha) * \hat{y}_{t-1}
\end{equation}

\begin{figure}[t]
    \centering
	\includegraphics[width=\columnwidth]{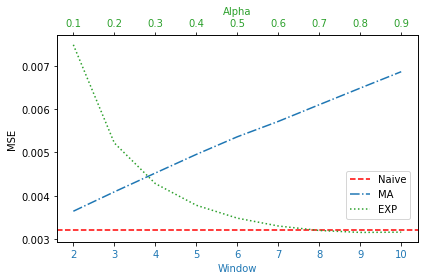}
	\caption{Naive, moving average, and exponential smoothing prediction MSE for varying hyperparameters.}
	\label{fig:classic}
\end{figure}

We are aware that the choice of window size and $\alpha$ significantly affect the prediction performance of the moving average and exponential smoothing methods. Figure~\ref{fig:classic}~shows the MA and EXP prediction MSE using different window sizes ($n$) and $\alpha$ values on the wind dataset. The plot shows that the wind dataset is highly complex, where naive almost entirely predicts better than MA or EXP (apart from a negligible improvement with EXP for $alpha$ values of 0.8 and 0.9). Due to this, we can only use the naive prediction MSE to represent classic prediction MSE in the following sections.

\begin{figure}[t]
    \centering
	\includegraphics[width=\columnwidth]{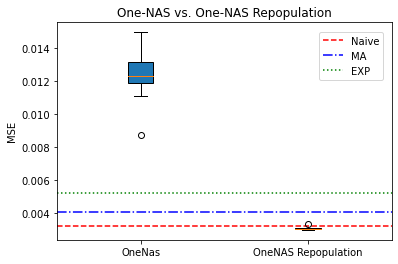}

	\caption{Online prediction MSE of ONE-NAS and One-NAS island repopulation}
	\label{fig:onenas-classic}
\end{figure}

Figure~\ref{fig:onenas-classic}~shows a box and whiskers plots for the the online prediction mean squared error (MSE) of ONE-NAS and ONE-NAS Repopulation algorithms over 10 repeated runs, alongside lines for each of the three classical time series forecasting methods (these methods are not randomized so their performance is always the same). Moving average (MA) prediction was done with a window size $n = 3$ and exponential smoothing was done with an $\alpha = 0.2$. The ONE-NAS Repopulation strategy shown in this plot used 20 islands (each with $5$ elite genomes and $10$ others per generation), with an extinction and repopulation frequency of 200 generations.

While the ONE-NAS online prediction performs worse than all three classical methods, the ONE-NAS Repopulation method not only significantly outperforms the base model, but has better online prediction performance than the classical methods across all repeats except for one outlier.

\subsection{One-NAS Repopulation}

\begin{figure}[t]
    \centering
	\includegraphics[width=\columnwidth]{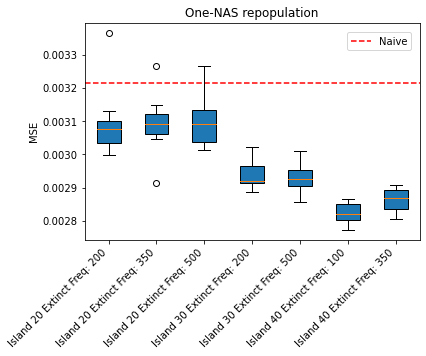}
	\caption{ONE-NAS evolution process}
	\label{fig:onenas_island}
\end{figure}

Given this encouragement that it is possible to effectively evolve and train RNNs in an online setting, we found that two hyperparameters significantly affect online prediction performance: island size and the extinction and repopulation frequency. 

Figure~\ref{fig:onenas_island}~shows a box plot of the online prediction MSE using island sizes of $20$, $30$, and $40$ over 10 repeated experiments with varying repopulation frequencies. As the number of islands increase, the prediction performance improves. Additionally, more frequent island repopulation also shows improvements in prediction performance, although not as significant. More islands allows more variety of species which allows the algorithm to have more chance to get out of the local optima, and also potentially provides more robustness to noise and overfitting of the data.

With same number of islands, more frequent extinction and repopulation events has better performance on average. Given that the number of generations is fixed at 2000 due to the length of the wind time series, an extinction frequency of 200 means at every 200 generations, the worst performing island is erased and repopulated, which results in 10 extinction and repopulation events. Once an island is erased, it is then repopulated with mutations of the current global best genome. So while this brings more variety to the population, the repopulated island needs time to evolve and eventually adapt to the environment. So the number of islands and total number of generations needs to be taken into consideration when choosing extinction and repopulation frequencies.

\subsection{Online Predictions over Time}

The previous plots show the overall performance of ONE-NAS and the classical methods for the entire wind time series, which in some sense provides an additional advantage to the classical methods in that they do not require any training, and as such have a significant advantage for earlier time steps when ONE-NAS has not had much opportunity to train and evolve RNNs.

\begin{figure}[t]
    \centering
	\includegraphics[width=\columnwidth]{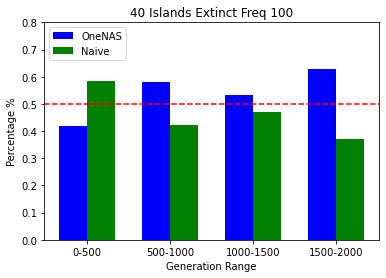}

	\caption{ONE-NAS percentage of prediction generations}
	\label{fig:40_islands_100}
\end{figure}

In order to investigate how much the RNNs evolved and trained by ONE-NAS were improving as they saw more data in an online fashion, we measured the for each generation how many time steps had a better prediction between the naive method and ONE-NAS. Figure~\ref{fig:40_islands_100}~shows the percentage of predictions that each method provided a more accurate prediction as the search progresses, with the red line representing 50\%.

From this plot we can observe that while ONE-NAS does not out perform the naive strategy within the first 500 generations, the performance of ONE-NAS continues to increase as the evolution progresses, which means ONE-NAS does not only train and predict values online, it also gets better throughout the evolutionary process which is what we would expect from an online algorithm. Also, it would be trivial to combine the two strategies, using a naive predictor until ONE-NAS has had enough evolution time to be more accurate.

However, we also notice that during generations 1500-2000, there are still more than 30\% of the generations that naive does better. In practice the sensor readings are not very accurate and sometimes some sensor readings do not change or have readings of $0$ for a while (\ie, they are noisy or sometimes turn off). Those situations are inevitable in practice and that data would be fed to an online predictor. When it happens, naive predictor can outperform ONE-NAS for periods of time while ONE-NAS readjusts. We did not remove invalid datapoints from the wind dataset for these experiments and we are presenting the raw prediction counts between naive and ONE-NAS, which further highlights its robustness on real-world data.

\subsection{Online Prediction Time Efficiency}

Another key concern for evaluation of online algorithms apart from prediction accuracy is time efficiency.  If an online algorithm cannot provide predictions at a rate less than the arrival rate of new data to be predicted than it is not usable. ONE-NAS somewhat alleviates this problem in that it utilizes the best genome from the previous generation to provide predictions while concurrently training the next generation.

\begin{table}[h]
    \centering
    \begin{tabular}{c c c} \hline

Num Islands & Avg Time (s) & Longest Time (s) \\ \hline \hline
        20 & 35.67 & 109.20 \\ \hline
        30 & 41.72 & 127.44  \\ \hline
        40 & 69.96 & 189.72  \\ \hline

    \end{tabular}
    \caption{\label{table:time} The average and longest time needed to train and evolve each generation for the wind dataset. 
}
\end{table}



Even so, for this work, each generation was generated and trained during a single subsequence of 25 time steps.  For the wind dataset, each time step was gathered at a 10 minute interval, so this provides a significant buffer, however for many time series datasets, time step frequency can be per minute, per second, or even faster - making time efficiency a serious concern. Table~\ref{table:time} presents the average and longest time required to evolve and train one generation of the ONE-NAS Repopulation experiments for the varying numbers of islands.  Note that population size was tied to island size, with $5$ elite genomes and $10$ other genomes per island, which is why the $40$ island genomes took approximately twice the time.  In the worst case for $40$ islands, the longest time per generation was a bit above $3$ minutes, which is far below the $250$ minute generations time for the wind data.  Variability for the generation evolution and training times comes both from the stochastic nature of how the RNN architectures are evolved, but also from the fact that the strategy will progressively grow larger RNNs, until a size is reached where adding additional components doesn't improve performance.

It should also be note that the workers training the genomes for each generation were distributed across 16 processors, and that performance will linearly scale upwards until the number of available processors was equal to the population size (i.e., all generated genomes can be independently trained in parallel without reduction in performance apart from fixed communication and genome generation overhead). This scalability of ONE-NAS makes it well suited to online learning.  For example, if we scaled up to $200$ processors from the $16$ used for the $20$ island experiments, we can estimate approximately $2.85$ seconds per generation (\ie, training the 200 non-elite genomes at once, instead of 16 at a time), plus some additional communication and generation overhead.  With a generation time of 25 time steps, this would allow for incoming data to be processed at almost $10$ readings each second. Given some flexibility in determining subsequence/generation time, ONE-NAS shows the potential to be able to operate for very high frequency time series given enough computing power.

\subsection{ONE-NAS vs Online ARIMA}
\label{sec:online_arima}

\begin{figure}[t]
    \centering
	\includegraphics[width=\columnwidth]{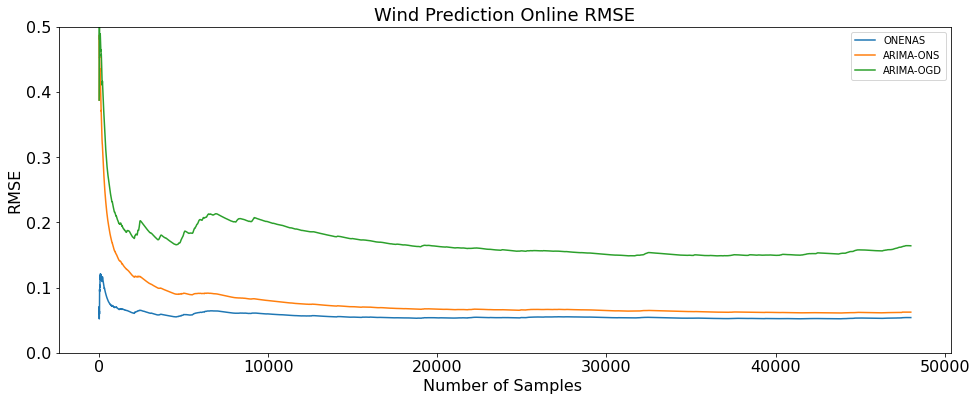}

	\caption{Online ARIMA vs. ONE-NAS predictions on the wind dataset. The plot shows how the online root mean squared error (RMSE) changes over time as more predictions are made.}
	\label{fig:wind_arima}
\end{figure}

\begin{figure}[t]
    \centering
	\includegraphics[width=\columnwidth]{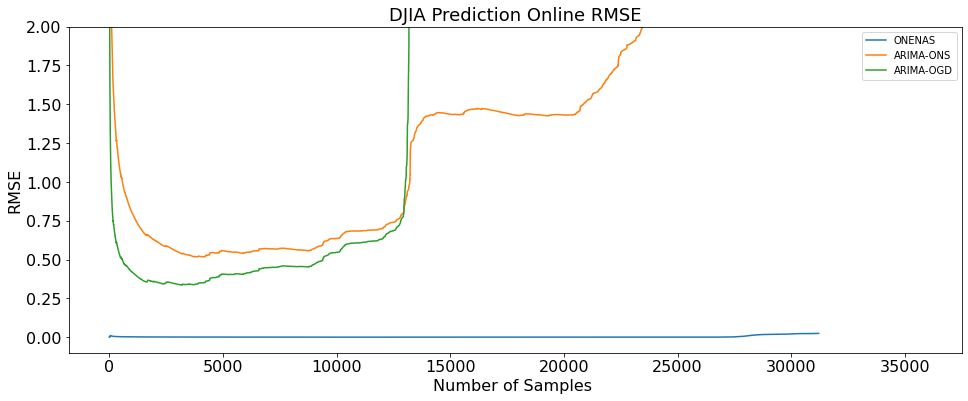}

	\caption{Online ARIMA vs. ONE-NAS predictions on the DJIA dataset. The plot shows how the online root mean squared error (RMSE) changes over time as more predictions are made.}
	\label{fig:djia_arima}
\end{figure}



While there is a significant lack of methods for online multivariate TSF, recent work by Liu \etal~\cite{liu2016online} has developed an online ARMIA method for univariate TSF. To compare ONE-NAS with a state of the art method as opposed to the classical methods investigated prior, we also compare ONE-NAS to this online ARIMA.

To reproduce results from Liu \etal, we first performed experiments using the Dow Jones Industrial Dataset (DJIA), which was used for evaluation in their work. Figure~\ref{fig:djia_arima} presents results for their ARIMA-ONS (Arima Online Newton Step) and ARIMA-OGD (ARIMA Online Gradient descent) variants, which were the best performing in their paper. These are compared to the ONE-NAS Repopulation method with 10 islands and an extinction frequency of 200, and a similar generation and subsequence length of $25$. For both ONE-NAS and online ARMIA, the plots show the online root mean squared error (RMSE) over time, from the average over 10 repeated experiments. The online RMSE over time is calculated as the average RSME of all previous predictions. For this DJIA data, we show that while the online ARIMA mirror results from their work, ONE-NAS outperforms this method by multiple orders of magnitude.

We then further compared ONE-NAS with online ARIMA on the wind datasets, which is shown in Figure~\ref{fig:wind_arima}. Similarly, the results show the average performance over 10 repeated experiments, however in this case ONE-NAS used the best hyperparameters from previous results (40 islands and an extinction frequency of 100). We performed a hyperparameter sweep for the online ARIMA methods, and selected the best hyperparameters. For ARIMA-ONS, the learning rate was set to $e^{-3}$ and $\epsilon = 3.16e^{-6}$. For ARIMA-OGD, learning rate was set to $e^3$, and $\epsilon = e^{-5.5}$. Similarly, ONE-NAS also outperforms the online ARIMA methods on the wind dataset.

%% file: 06-conclusion.tex
\section{Conclusion}
\label{sec:conclusion}

This work presents our novel Online NeuroEvolution based Neural Architecture Search (ONE-NAS) algorithm and applies it to time series forecasting (TSF) on a challenging real world wind turbine dataset. To the authors knowledge, ONE-NAS is the first neural architecture search algorithm capable of designing and training recurrent neural networks in real time as data arrives in an online scenario. We show that after an initial burn-in learning period, ONE-NAS outperforms traditional online statistical time series forecasting methods such as naive, moving average and exponential smoothing, as well as modern online ARMIA methods. ONE-NAS uses a novel strategy where generations of RNNs are evolved for TSF while concurrently the best RNN from the previous generation is used for generating predictions. The method is distributed and highly scalable to compute resources, allowing for generated RNNs to be trained in parallel, and performance results show that the algorithm is capable of operating on high frequency (\eg, per second) data streams given enough compute resources.

One major advancement which allowed ONE-NAS to be capable of accurate online predictions was the use of multiple islands of populations, where at specific frequencies the worst performing islands are deleted and repopulated with mutations of the global best genome (RNN). While the single population ONE-NAS had poor performance, we found that increasing the number of islands provided significant increases in performance, and also that increasing the extinction frequency also improved performance, albiet less dramatically.  This strategy helps increase the diversity of the RNNs across islands, allowing for a more robust population which can also break out of local optima.

ONE-NAS eliminates the need to perform offline retraining or redesigning of RNNs for new time series data, and shows good results on a real world, noisy dataset which includes periods of time where various sensors are offline. It also opens the door to significant future work, as it shows that the development of online neuroevolution algorithms is possible, as they can easily take advantage of distributed and parallel computing strategies.

\section{Future Work}
\label{sec:future_work}

The ability to dynamically evolve and train populations of recurrent neural networks in online scenarios, which can outperform traditional methods, is very exciting and opens up significant avenues for future work.  ONE-NAS as presented has a number of opportunities for further increasing performance. Currently, ONE-NAS continually increases its historical training data, however being selective about what historical data to retain and use for training could stand to benefit performance. Additionally, ONE-NAS currently uses a fixed generation time, it may be possible to dynamically adjust this to find more optimal generations that better align with computational requirements for real time predictions.

Additionally, while we provided a brief discussion of ONE-NAS's performance, doing a serious performance and scalability study could help inform and find areas for futher performance optimizations to improve its how quickly it can generate, train and perform predictions. We further plan to investigate ONE-NAS on other datasets, as well as compare it to fixed architecture RNN strategies~\cite{guo2016robust,yang2017robust,wang2014online}.

Finally, and perhaps most interesting, ONE-NAS provides populations of RNNs which can potentially provide online predictions at the end of each generation, and current results are based on utilizing only the RNN which performed best on the selected validation set. It may be possible to utilize selected RNNs as ensembles to further improve predictive ability, or even provide confidence or error bounds as to the potential accuracy of the best RNNs predictions. It is also possible to investigate retaining RNNs that perform well on historical data to further reduce or prevent catastrophic forgetting.

%% file: 00-main.bbl

\begin{thebibliography}{47}


\ifx \showCODEN    \undefined \def \showCODEN     #1{\unskip}     \fi
\ifx \showDOI      \undefined \def \showDOI       #1{#1}\fi
\ifx \showISBNx    \undefined \def \showISBNx     #1{\unskip}     \fi
\ifx \showISBNxiii \undefined \def \showISBNxiii  #1{\unskip}     \fi
\ifx \showISSN     \undefined \def \showISSN      #1{\unskip}     \fi
\ifx \showLCCN     \undefined \def \showLCCN      #1{\unskip}     \fi
\ifx \shownote     \undefined \def \shownote      #1{#1}          \fi
\ifx \showarticletitle \undefined \def \showarticletitle #1{#1}   \fi
\ifx \showURL      \undefined \def \showURL       {\relax}        \fi
\providecommand\bibfield[2]{#2}
\providecommand\bibinfo[2]{#2}
\providecommand\natexlab[1]{#1}
\providecommand\showeprint[2][]{arXiv:#2}

\bibitem[\protect\citeauthoryear{Abraham and Robins}{Abraham and
  Robins}{2005}]%
        {abraham2005memory}
\bibfield{author}{\bibinfo{person}{Wickliffe~C Abraham} {and}
  \bibinfo{person}{Anthony Robins}.} \bibinfo{year}{2005}\natexlab{}.
\newblock \showarticletitle{Memory retention--the synaptic stability versus
  plasticity dilemma}.
\newblock \bibinfo{journal}{\emph{Trends in neurosciences}}
  \bibinfo{volume}{28}, \bibinfo{number}{2} (\bibinfo{year}{2005}),
  \bibinfo{pages}{73--78}.
\newblock


\bibitem[\protect\citeauthoryear{Agogino, Stanley, and Miikkulainen}{Agogino
  et~al\mbox{.}}{2000}]%
        {agogino2000online}
\bibfield{author}{\bibinfo{person}{Adrian Agogino}, \bibinfo{person}{Kenneth
  Stanley}, {and} \bibinfo{person}{Risto Miikkulainen}.}
  \bibinfo{year}{2000}\natexlab{}.
\newblock \showarticletitle{Online interactive neuro-evolution}.
\newblock \bibinfo{journal}{\emph{Neural Processing Letters}}
  \bibinfo{volume}{11}, \bibinfo{number}{1} (\bibinfo{year}{2000}),
  \bibinfo{pages}{29--38}.
\newblock


\bibitem[\protect\citeauthoryear{Al-Musaylh, Deo, and Li}{Al-Musaylh
  et~al\mbox{.}}{2020}]%
        {al2020electrical}
\bibfield{author}{\bibinfo{person}{Mohanad~S Al-Musaylh},
  \bibinfo{person}{Ravinesh~C Deo}, {and} \bibinfo{person}{Yan Li}.}
  \bibinfo{year}{2020}\natexlab{}.
\newblock \showarticletitle{Electrical energy demand forecasting model
  development and evaluation with maximum overlap discrete wavelet
  transform-online sequential extreme learning machines algorithms}.
\newblock \bibinfo{journal}{\emph{Energies}} \bibinfo{volume}{13},
  \bibinfo{number}{9} (\bibinfo{year}{2020}), \bibinfo{pages}{2307}.
\newblock


\bibitem[\protect\citeauthoryear{Anava, Hazan, Mannor, and Shamir}{Anava
  et~al\mbox{.}}{2013}]%
        {anava2013online}
\bibfield{author}{\bibinfo{person}{Oren Anava}, \bibinfo{person}{Elad Hazan},
  \bibinfo{person}{Shie Mannor}, {and} \bibinfo{person}{Ohad Shamir}.}
  \bibinfo{year}{2013}\natexlab{}.
\newblock \showarticletitle{Online learning for time series prediction}. In
  \bibinfo{booktitle}{\emph{Conference on learning theory}}. PMLR,
  \bibinfo{pages}{172--184}.
\newblock


\bibitem[\protect\citeauthoryear{Anava, Hazan, and Zeevi}{Anava
  et~al\mbox{.}}{2015}]%
        {anava2015online}
\bibfield{author}{\bibinfo{person}{Oren Anava}, \bibinfo{person}{Elad Hazan},
  {and} \bibinfo{person}{Assaf Zeevi}.} \bibinfo{year}{2015}\natexlab{}.
\newblock \showarticletitle{Online time series prediction with missing data}.
  In \bibinfo{booktitle}{\emph{International Conference on Machine Learning}}.
  PMLR, \bibinfo{pages}{2191--2199}.
\newblock


\bibitem[\protect\citeauthoryear{Bhatnagar, Lal, Gupta, Gupta,
  et~al\mbox{.}}{Bhatnagar et~al\mbox{.}}{2012}]%
        {bhatnagar2012forecasting}
\bibfield{author}{\bibinfo{person}{Sunil Bhatnagar}, \bibinfo{person}{Vivek
  Lal}, \bibinfo{person}{Shiv~D Gupta}, \bibinfo{person}{Om~P Gupta},
  {et~al\mbox{.}}} \bibinfo{year}{2012}\natexlab{}.
\newblock \showarticletitle{Forecasting incidence of dengue in Rajasthan, using
  time series analyses}.
\newblock \bibinfo{journal}{\emph{Indian journal of public health}}
  \bibinfo{volume}{56}, \bibinfo{number}{4} (\bibinfo{year}{2012}),
  \bibinfo{pages}{281}.
\newblock


\bibitem[\protect\citeauthoryear{Bishop}{Bishop}{1995}]%
        {bishop1995training}
\bibfield{author}{\bibinfo{person}{Chris~M Bishop}.}
  \bibinfo{year}{1995}\natexlab{}.
\newblock \showarticletitle{Training with noise is equivalent to Tikhonov
  regularization}.
\newblock \bibinfo{journal}{\emph{Neural computation}} \bibinfo{volume}{7},
  \bibinfo{number}{1} (\bibinfo{year}{1995}), \bibinfo{pages}{108--116}.
\newblock


\bibitem[\protect\citeauthoryear{Bishop et~al\mbox{.}}{Bishop
  et~al\mbox{.}}{1995}]%
        {bishop1995neural}
\bibfield{author}{\bibinfo{person}{Christopher~M Bishop} {et~al\mbox{.}}}
  \bibinfo{year}{1995}\natexlab{}.
\newblock \bibinfo{booktitle}{\emph{Neural networks for pattern recognition}}.
\newblock \bibinfo{publisher}{Oxford university press}.
\newblock


\bibitem[\protect\citeauthoryear{Blumofe and Leiserson}{Blumofe and
  Leiserson}{1994}]%
        {mit-cilk-94}
\bibfield{author}{\bibinfo{person}{R.~D. Blumofe} {and} \bibinfo{person}{C.~E.
  Leiserson}.} \bibinfo{year}{1994}\natexlab{}.
\newblock \showarticletitle{{Scheduling Multithreaded Computations by Work
  Stealing}}. In \bibinfo{booktitle}{\emph{Proceedings of the 35th Annual
  Symposium on Foundations of Computer Science (FOCS '94)}}.
  \bibinfo{address}{Santa Fe, New Mexico}, \bibinfo{pages}{356--368}.
\newblock


\bibitem[\protect\citeauthoryear{Cao, Li, and Li}{Cao et~al\mbox{.}}{2019}]%
        {cao2019financial}
\bibfield{author}{\bibinfo{person}{Jian Cao}, \bibinfo{person}{Zhi Li}, {and}
  \bibinfo{person}{Jian Li}.} \bibinfo{year}{2019}\natexlab{}.
\newblock \showarticletitle{Financial time series forecasting model based on
  CEEMDAN and LSTM}.
\newblock \bibinfo{journal}{\emph{Physica A: Statistical Mechanics and its
  Applications}}  \bibinfo{volume}{519} (\bibinfo{year}{2019}),
  \bibinfo{pages}{127--139}.
\newblock


\bibitem[\protect\citeauthoryear{Cardamone, Loiacono, and Lanzi}{Cardamone
  et~al\mbox{.}}{2010}]%
        {cardamone2010learning}
\bibfield{author}{\bibinfo{person}{Luigi Cardamone}, \bibinfo{person}{Daniele
  Loiacono}, {and} \bibinfo{person}{Pier~Luca Lanzi}.}
  \bibinfo{year}{2010}\natexlab{}.
\newblock \showarticletitle{Learning to drive in the open racing car simulator
  using online neuroevolution}.
\newblock \bibinfo{journal}{\emph{IEEE Transactions on Computational
  Intelligence and AI in Games}} \bibinfo{volume}{2}, \bibinfo{number}{3}
  (\bibinfo{year}{2010}), \bibinfo{pages}{176--190}.
\newblock


\bibitem[\protect\citeauthoryear{Chung, Gulcehre, Cho, and Bengio}{Chung
  et~al\mbox{.}}{2014}]%
        {chung2014empirical}
\bibfield{author}{\bibinfo{person}{Junyoung Chung}, \bibinfo{person}{Caglar
  Gulcehre}, \bibinfo{person}{KyungHyun Cho}, {and} \bibinfo{person}{Yoshua
  Bengio}.} \bibinfo{year}{2014}\natexlab{}.
\newblock \showarticletitle{Empirical evaluation of gated recurrent neural
  networks on sequence modeling}.
\newblock \bibinfo{journal}{\emph{arXiv preprint arXiv:1412.3555}}
  (\bibinfo{year}{2014}).
\newblock


\bibitem[\protect\citeauthoryear{Collins, Sohl-Dickstein, and Sussillo}{Collins
  et~al\mbox{.}}{2016}]%
        {collins2016capacity}
\bibfield{author}{\bibinfo{person}{Jasmine Collins}, \bibinfo{person}{Jascha
  Sohl-Dickstein}, {and} \bibinfo{person}{David Sussillo}.}
  \bibinfo{year}{2016}\natexlab{}.
\newblock \showarticletitle{Capacity and Trainability in Recurrent Neural
  Networks}.
\newblock \bibinfo{journal}{\emph{arXiv preprint arXiv:1611.09913}}
  (\bibinfo{year}{2016}).
\newblock


\bibitem[\protect\citeauthoryear{ElSaid, Karnas, Lyu, Krutz, Ororbia, and
  Desell}{ElSaid et~al\mbox{.}}{2020a}]%
        {elsaid2020neuro}
\bibfield{author}{\bibinfo{person}{AbdElRahman ElSaid}, \bibinfo{person}{Joshua
  Karnas}, \bibinfo{person}{Zimeng Lyu}, \bibinfo{person}{Daniel Krutz},
  \bibinfo{person}{Alexander~G Ororbia}, {and} \bibinfo{person}{Travis
  Desell}.} \bibinfo{year}{2020}\natexlab{a}.
\newblock \showarticletitle{Neuro-Evolutionary Transfer Learning through
  Structural Adaptation}. In \bibinfo{booktitle}{\emph{International Conference
  on the Applications of Evolutionary Computation (Part of EvoStar)}}.
  Springer, \bibinfo{pages}{610--625}.
\newblock


\bibitem[\protect\citeauthoryear{ElSaid, Karns, Lyu, Krutz, Ororbia, and
  Desell}{ElSaid et~al\mbox{.}}{2020b}]%
        {elsaid2020improving}
\bibfield{author}{\bibinfo{person}{AbdElRahman ElSaid}, \bibinfo{person}{Joshua
  Karns}, \bibinfo{person}{Zimeng Lyu}, \bibinfo{person}{Daniel Krutz},
  \bibinfo{person}{Alexander Ororbia}, {and} \bibinfo{person}{Travis Desell}.}
  \bibinfo{year}{2020}\natexlab{b}.
\newblock \showarticletitle{Improving neuroevolutionary transfer learning of
  deep recurrent neural networks through network-aware adaptation}. In
  \bibinfo{booktitle}{\emph{Proceedings of the 2020 Genetic and Evolutionary
  Computation Conference}}. \bibinfo{pages}{315--323}.
\newblock


\bibitem[\protect\citeauthoryear{Galassi, Capodieci, Cabri, and
  Leonardi}{Galassi et~al\mbox{.}}{2016}]%
        {galassi2016evolutionary}
\bibfield{author}{\bibinfo{person}{Marco Galassi}, \bibinfo{person}{Nicola
  Capodieci}, \bibinfo{person}{Giacomo Cabri}, {and} \bibinfo{person}{Letizia
  Leonardi}.} \bibinfo{year}{2016}\natexlab{}.
\newblock \showarticletitle{Evolutionary strategies for novelty-based online
  neuroevolution in swarm robotics}. In \bibinfo{booktitle}{\emph{2016 IEEE
  International Conference on Systems, Man, and Cybernetics (SMC)}}. IEEE,
  \bibinfo{pages}{002026--002032}.
\newblock


\bibitem[\protect\citeauthoryear{Ghosh, Basu, and O’Mahony}{Ghosh
  et~al\mbox{.}}{2005}]%
        {ghosh2005time}
\bibfield{author}{\bibinfo{person}{Bidisha Ghosh}, \bibinfo{person}{Biswajit
  Basu}, {and} \bibinfo{person}{Margaret O’Mahony}.}
  \bibinfo{year}{2005}\natexlab{}.
\newblock \showarticletitle{Time-series modelling for forecasting vehicular
  traffic flow in Dublin}. In \bibinfo{booktitle}{\emph{84th Annual Meeting of
  the Transportation Research Board, Washington, DC}}.
\newblock


\bibitem[\protect\citeauthoryear{Guo, Xu, Yao, Chen, Aberer, and Funaya}{Guo
  et~al\mbox{.}}{2016}]%
        {guo2016robust}
\bibfield{author}{\bibinfo{person}{Tian Guo}, \bibinfo{person}{Zhao Xu},
  \bibinfo{person}{Xin Yao}, \bibinfo{person}{Haifeng Chen},
  \bibinfo{person}{Karl Aberer}, {and} \bibinfo{person}{Koichi Funaya}.}
  \bibinfo{year}{2016}\natexlab{}.
\newblock \showarticletitle{Robust online time series prediction with recurrent
  neural networks}. In \bibinfo{booktitle}{\emph{2016 IEEE International
  Conference on Data Science and Advanced Analytics (DSAA)}}. Ieee,
  \bibinfo{pages}{816--825}.
\newblock


\bibitem[\protect\citeauthoryear{Han, Zhang, Xu, Qiu, and Wang}{Han
  et~al\mbox{.}}{2018}]%
        {han2018multivariate}
\bibfield{author}{\bibinfo{person}{Min Han}, \bibinfo{person}{Shuhui Zhang},
  \bibinfo{person}{Meiling Xu}, \bibinfo{person}{Tie Qiu}, {and}
  \bibinfo{person}{Ning Wang}.} \bibinfo{year}{2018}\natexlab{}.
\newblock \showarticletitle{Multivariate chaotic time series online prediction
  based on improved kernel recursive least squares algorithm}.
\newblock \bibinfo{journal}{\emph{IEEE transactions on cybernetics}}
  \bibinfo{volume}{49}, \bibinfo{number}{4} (\bibinfo{year}{2018}),
  \bibinfo{pages}{1160--1172}.
\newblock


\bibitem[\protect\citeauthoryear{Hochreiter and Schmidhuber}{Hochreiter and
  Schmidhuber}{1997}]%
        {hochreiter1997long}
\bibfield{author}{\bibinfo{person}{Sepp Hochreiter} {and}
  \bibinfo{person}{J{\"u}rgen Schmidhuber}.} \bibinfo{year}{1997}\natexlab{}.
\newblock \showarticletitle{Long short-term memory}.
\newblock \bibinfo{journal}{\emph{Neural Computation}} \bibinfo{volume}{9},
  \bibinfo{number}{8} (\bibinfo{year}{1997}), \bibinfo{pages}{1735--1780}.
\newblock


\bibitem[\protect\citeauthoryear{Jozefowicz, Zaremba, and Sutskever}{Jozefowicz
  et~al\mbox{.}}{2015}]%
        {jozefowicz2015empirical}
\bibfield{author}{\bibinfo{person}{Rafal Jozefowicz}, \bibinfo{person}{Wojciech
  Zaremba}, {and} \bibinfo{person}{Ilya Sutskever}.}
  \bibinfo{year}{2015}\natexlab{}.
\newblock \showarticletitle{An empirical exploration of recurrent network
  architectures}. In \bibinfo{booktitle}{\emph{International Conference on
  Machine Learning}}. \bibinfo{pages}{2342--2350}.
\newblock


\bibitem[\protect\citeauthoryear{Kozitsin, Katser, and Lakontsev}{Kozitsin
  et~al\mbox{.}}{2021}]%
        {kozitsin2021online}
\bibfield{author}{\bibinfo{person}{Viacheslav Kozitsin}, \bibinfo{person}{Iurii
  Katser}, {and} \bibinfo{person}{Dmitry Lakontsev}.}
  \bibinfo{year}{2021}\natexlab{}.
\newblock \showarticletitle{Online Forecasting and Anomaly Detection Based on
  the ARIMA Model}.
\newblock \bibinfo{journal}{\emph{Applied Sciences}} \bibinfo{volume}{11},
  \bibinfo{number}{7} (\bibinfo{year}{2021}), \bibinfo{pages}{3194}.
\newblock


\bibitem[\protect\citeauthoryear{Lim and Zohren}{Lim and Zohren}{2021}]%
        {lim2021time}
\bibfield{author}{\bibinfo{person}{Bryan Lim} {and} \bibinfo{person}{Stefan
  Zohren}.} \bibinfo{year}{2021}\natexlab{}.
\newblock \showarticletitle{Time-series forecasting with deep learning: a
  survey}.
\newblock \bibinfo{journal}{\emph{Philosophical Transactions of the Royal
  Society A}} \bibinfo{volume}{379}, \bibinfo{number}{2194}
  (\bibinfo{year}{2021}), \bibinfo{pages}{20200209}.
\newblock


\bibitem[\protect\citeauthoryear{Liu, Hoi, Zhao, and Sun}{Liu
  et~al\mbox{.}}{2016}]%
        {liu2016online}
\bibfield{author}{\bibinfo{person}{Chenghao Liu}, \bibinfo{person}{Steven~CH
  Hoi}, \bibinfo{person}{Peilin Zhao}, {and} \bibinfo{person}{Jianling Sun}.}
  \bibinfo{year}{2016}\natexlab{}.
\newblock \showarticletitle{Online arima algorithms for time series
  prediction}. In \bibinfo{booktitle}{\emph{Thirtieth AAAI conference on
  artificial intelligence}}.
\newblock


\bibitem[\protect\citeauthoryear{Lyu, ElSaid, Karns, Mkaouer, and Desell}{Lyu
  et~al\mbox{.}}{2021a}]%
        {lyu2021experimental}
\bibfield{author}{\bibinfo{person}{Zimeng Lyu}, \bibinfo{person}{AbdElRahman
  ElSaid}, \bibinfo{person}{Joshua Karns}, \bibinfo{person}{Mohamed Mkaouer},
  {and} \bibinfo{person}{Travis Desell}.} \bibinfo{year}{2021}\natexlab{a}.
\newblock \showarticletitle{An Experimental Study of Weight Initialization and
  Lamarckian Inheritance on Neuroevolution}.
\newblock \bibinfo{journal}{\emph{The 24th International Conference on the
  Applications of Evolutionary Computation (EvoStar: EvoApps)}}
  (\bibinfo{year}{2021}).
\newblock


\bibitem[\protect\citeauthoryear{Lyu, Karnas, ElSaid, Mkaouer, and Desell}{Lyu
  et~al\mbox{.}}{2021b}]%
        {lyu2021Improving}
\bibfield{author}{\bibinfo{person}{Zimeng Lyu}, \bibinfo{person}{Joshua
  Karnas}, \bibinfo{person}{AbdElRahman ElSaid}, \bibinfo{person}{Mohamed
  Mkaouer}, {and} \bibinfo{person}{Travis Desell}.}
  \bibinfo{year}{2021}\natexlab{b}.
\newblock \showarticletitle{Improving Distributed Neuroevolution Using Island
  Extinction and Repopulation}.
\newblock \bibinfo{journal}{\emph{The 24th International Conference on the
  Applications of Evolutionary Computation (EvoStar: EvoApps)}}
  (\bibinfo{year}{2021}).
\newblock


\bibitem[\protect\citeauthoryear{Lyu, Patwardhan, Stadem, Langfeld, Benson,
  Thoelke, and Desell}{Lyu et~al\mbox{.}}{2021c}]%
        {lyu2021neuroevolution}
\bibfield{author}{\bibinfo{person}{Zimeng Lyu}, \bibinfo{person}{Shuchita
  Patwardhan}, \bibinfo{person}{David Stadem}, \bibinfo{person}{James
  Langfeld}, \bibinfo{person}{Steve Benson}, \bibinfo{person}{Seth Thoelke},
  {and} \bibinfo{person}{Travis Desell}.} \bibinfo{year}{2021}\natexlab{c}.
\newblock \showarticletitle{Neuroevolution of recurrent neural networks for
  time series forecasting of coal-fired power plant operating parameters}. In
  \bibinfo{booktitle}{\emph{Proceedings of the Genetic and Evolutionary
  Computation Conference Companion}}. \bibinfo{pages}{1735--1743}.
\newblock


\bibitem[\protect\citeauthoryear{Mahalakshmi, Sridevi, and Rajaram}{Mahalakshmi
  et~al\mbox{.}}{2016}]%
        {mahalakshmi2016survey}
\bibfield{author}{\bibinfo{person}{Ganapathy Mahalakshmi}, \bibinfo{person}{S
  Sridevi}, {and} \bibinfo{person}{Shyamsundar Rajaram}.}
  \bibinfo{year}{2016}\natexlab{}.
\newblock \showarticletitle{A survey on forecasting of time series data}. In
  \bibinfo{booktitle}{\emph{2016 International Conference on Computing
  Technologies and Intelligent Data Engineering (ICCTIDE'16)}}. IEEE,
  \bibinfo{pages}{1--8}.
\newblock


\bibitem[\protect\citeauthoryear{Makridakis, Spiliotis, and
  Assimakopoulos}{Makridakis et~al\mbox{.}}{2018}]%
        {makridakis2018statistical}
\bibfield{author}{\bibinfo{person}{Spyros Makridakis},
  \bibinfo{person}{Evangelos Spiliotis}, {and} \bibinfo{person}{Vassilios
  Assimakopoulos}.} \bibinfo{year}{2018}\natexlab{}.
\newblock \showarticletitle{Statistical and Machine Learning forecasting
  methods: Concerns and ways forward}.
\newblock \bibinfo{journal}{\emph{PloS one}} \bibinfo{volume}{13},
  \bibinfo{number}{3} (\bibinfo{year}{2018}), \bibinfo{pages}{e0194889}.
\newblock


\bibitem[\protect\citeauthoryear{McCloskey and Cohen}{McCloskey and
  Cohen}{1989}]%
        {mccloskey1989catastrophic}
\bibfield{author}{\bibinfo{person}{Michael McCloskey} {and}
  \bibinfo{person}{Neal~J Cohen}.} \bibinfo{year}{1989}\natexlab{}.
\newblock \showarticletitle{Catastrophic interference in connectionist
  networks: The sequential learning problem}.
\newblock In \bibinfo{booktitle}{\emph{Psychology of learning and motivation}}.
  Vol.~\bibinfo{volume}{24}. \bibinfo{publisher}{Elsevier},
  \bibinfo{pages}{109--165}.
\newblock


\bibitem[\protect\citeauthoryear{Ororbia, ElSaid, and Desell}{Ororbia
  et~al\mbox{.}}{2019}]%
        {ororbia2019examm}
\bibfield{author}{\bibinfo{person}{Alexander Ororbia},
  \bibinfo{person}{AbdElRahman ElSaid}, {and} \bibinfo{person}{Travis Desell}.}
  \bibinfo{year}{2019}\natexlab{}.
\newblock \showarticletitle{Investigating Recurrent Neural Network Memory
  Structures Using Neuro-evolution}. In \bibinfo{booktitle}{\emph{Proceedings
  of the Genetic and Evolutionary Computation Conference}} (Prague, Czech
  Republic) \emph{(\bibinfo{series}{GECCO '19})}. \bibinfo{publisher}{ACM},
  \bibinfo{address}{New York, NY, USA}, \bibinfo{pages}{446--455}.
\newblock
\showISBNx{978-1-4503-6111-8}
\urldef\tempurl%
\url{https://doi.org/10.1145/3321707.3321795}
\showDOI{\tempurl}


\bibitem[\protect\citeauthoryear{Ororbia, Alexander, Linder, and Snoke}{Ororbia
  et~al\mbox{.}}{2018}]%
        {ororbia2018using}
\bibfield{author}{\bibinfo{person}{II Ororbia}, \bibinfo{person}{G Alexander},
  \bibinfo{person}{Fridolin Linder}, {and} \bibinfo{person}{Joshua Snoke}.}
  \bibinfo{year}{2018}\natexlab{}.
\newblock \showarticletitle{Using Neural Generative Models to Release Synthetic
  Twitter Corpora with Reduced Stylometric Identifiability of Users}.
\newblock \bibinfo{journal}{\emph{arXiv preprint arXiv:1606.01151}}
  (\bibinfo{year}{2018}).
\newblock


\bibitem[\protect\citeauthoryear{Ororbia~II, Mikolov, and Reitter}{Ororbia~II
  et~al\mbox{.}}{2017}]%
        {ororbia2017diff}
\bibfield{author}{\bibinfo{person}{Alexander~G. Ororbia~II},
  \bibinfo{person}{Tomas Mikolov}, {and} \bibinfo{person}{David Reitter}.}
  \bibinfo{year}{2017}\natexlab{}.
\newblock \showarticletitle{Learning Simpler Language Models with the
  Differential State Framework}.
\newblock \bibinfo{journal}{\emph{Neural Computation}} \bibinfo{volume}{0},
  \bibinfo{number}{0} (\bibinfo{year}{2017}), \bibinfo{pages}{1--26}.
\newblock
\urldef\tempurl%
\url{https://doi.org/10.1162/neco\_a\_01017}
\showDOI{\tempurl}
\showeprint{https://doi.org/10.1162/neco\_a\_01017}
\newblock
\shownote{PMID: 28957029}.


\bibitem[\protect\citeauthoryear{Osegi}{Osegi}{2020}]%
        {osegi2020using}
\bibfield{author}{\bibinfo{person}{EN Osegi}.} \bibinfo{year}{2020}\natexlab{}.
\newblock \showarticletitle{Using the hierarchical temporal memory spatial
  pooler for short-term forecasting of electrical load time series}.
\newblock \bibinfo{journal}{\emph{Applied Computing and Informatics}}
  (\bibinfo{year}{2020}).
\newblock


\bibitem[\protect\citeauthoryear{Pascanu, Mikolov, and Bengio}{Pascanu
  et~al\mbox{.}}{2013}]%
        {pascanu2013difficulty}
\bibfield{author}{\bibinfo{person}{Razvan Pascanu}, \bibinfo{person}{Tomas
  Mikolov}, {and} \bibinfo{person}{Yoshua Bengio}.}
  \bibinfo{year}{2013}\natexlab{}.
\newblock \showarticletitle{On the difficulty of training recurrent neural
  networks}. In \bibinfo{booktitle}{\emph{International Conference on Machine
  Learning}}. \bibinfo{pages}{1310--1318}.
\newblock


\bibitem[\protect\citeauthoryear{Sahoo, Pham, Lu, and Hoi}{Sahoo
  et~al\mbox{.}}{2017}]%
        {sahoo2017online}
\bibfield{author}{\bibinfo{person}{Doyen Sahoo}, \bibinfo{person}{Quang Pham},
  \bibinfo{person}{Jing Lu}, {and} \bibinfo{person}{Steven~CH Hoi}.}
  \bibinfo{year}{2017}\natexlab{}.
\newblock \showarticletitle{Online deep learning: Learning deep neural networks
  on the fly}.
\newblock \bibinfo{journal}{\emph{arXiv preprint arXiv:1711.03705}}
  (\bibinfo{year}{2017}).
\newblock


\bibitem[\protect\citeauthoryear{Schmidt, Suri-Payer, Gulenko,
  Wallschl{\"a}ger, Acker, and Kao}{Schmidt et~al\mbox{.}}{2018}]%
        {schmidt2018unsupervised}
\bibfield{author}{\bibinfo{person}{Florian Schmidt}, \bibinfo{person}{Florian
  Suri-Payer}, \bibinfo{person}{Anton Gulenko}, \bibinfo{person}{Marcel
  Wallschl{\"a}ger}, \bibinfo{person}{Alexander Acker}, {and}
  \bibinfo{person}{Odej Kao}.} \bibinfo{year}{2018}\natexlab{}.
\newblock \showarticletitle{Unsupervised anomaly event detection for cloud
  monitoring using online arima}. In \bibinfo{booktitle}{\emph{2018 IEEE/ACM
  International Conference on Utility and Cloud Computing Companion (UCC
  Companion)}}. IEEE, \bibinfo{pages}{71--76}.
\newblock


\bibitem[\protect\citeauthoryear{Stanley and Miikkulainen}{Stanley and
  Miikkulainen}{2002}]%
        {stanley2002evolving}
\bibfield{author}{\bibinfo{person}{Kenneth Stanley} {and}
  \bibinfo{person}{Risto Miikkulainen}.} \bibinfo{year}{2002}\natexlab{}.
\newblock \showarticletitle{Evolving neural networks through augmenting
  topologies}.
\newblock \bibinfo{journal}{\emph{Evolutionary computation}}
  \bibinfo{volume}{10}, \bibinfo{number}{2} (\bibinfo{year}{2002}),
  \bibinfo{pages}{99--127}.
\newblock


\bibitem[\protect\citeauthoryear{Stanley, Bryant, and Miikkulainen}{Stanley
  et~al\mbox{.}}{2005}]%
        {stanley2005real}
\bibfield{author}{\bibinfo{person}{Kenneth~O Stanley}, \bibinfo{person}{Bobby~D
  Bryant}, {and} \bibinfo{person}{Risto Miikkulainen}.}
  \bibinfo{year}{2005}\natexlab{}.
\newblock \showarticletitle{Real-time neuroevolution in the NERO video game}.
\newblock \bibinfo{journal}{\emph{IEEE transactions on evolutionary
  computation}} \bibinfo{volume}{9}, \bibinfo{number}{6}
  (\bibinfo{year}{2005}), \bibinfo{pages}{653--668}.
\newblock


\bibitem[\protect\citeauthoryear{Sutton and Barto}{Sutton and Barto}{2018}]%
        {sutton2018reinforcement}
\bibfield{author}{\bibinfo{person}{Richard~S Sutton} {and}
  \bibinfo{person}{Andrew~G Barto}.} \bibinfo{year}{2018}\natexlab{}.
\newblock \bibinfo{booktitle}{\emph{Reinforcement learning: An introduction}}.
\newblock \bibinfo{publisher}{MIT press}.
\newblock


\bibitem[\protect\citeauthoryear{Torres, Hadjout, Sebaa,
  Mart{\'\i}nez-{\'A}lvarez, and Troncoso}{Torres et~al\mbox{.}}{2021}]%
        {torres2021deep}
\bibfield{author}{\bibinfo{person}{Jos{\'e}~F Torres}, \bibinfo{person}{Dalil
  Hadjout}, \bibinfo{person}{Abderrazak Sebaa}, \bibinfo{person}{Francisco
  Mart{\'\i}nez-{\'A}lvarez}, {and} \bibinfo{person}{Alicia Troncoso}.}
  \bibinfo{year}{2021}\natexlab{}.
\newblock \showarticletitle{Deep learning for time series forecasting: a
  survey}.
\newblock \bibinfo{journal}{\emph{Big Data}} \bibinfo{volume}{9},
  \bibinfo{number}{1} (\bibinfo{year}{2021}), \bibinfo{pages}{3--21}.
\newblock


\bibitem[\protect\citeauthoryear{Vlasenko, Vlasenko, Vynokurova, Bodyanskiy,
  and Peleshko}{Vlasenko et~al\mbox{.}}{2019}]%
        {vlasenko2019novel}
\bibfield{author}{\bibinfo{person}{Alexander Vlasenko},
  \bibinfo{person}{Nataliia Vlasenko}, \bibinfo{person}{Olena Vynokurova},
  \bibinfo{person}{Yevgeniy Bodyanskiy}, {and} \bibinfo{person}{Dmytro
  Peleshko}.} \bibinfo{year}{2019}\natexlab{}.
\newblock \showarticletitle{A Novel ensemble neuro-fuzzy model for financial
  time series forecasting}.
\newblock \bibinfo{journal}{\emph{Data}} \bibinfo{volume}{4},
  \bibinfo{number}{3} (\bibinfo{year}{2019}), \bibinfo{pages}{126}.
\newblock


\bibitem[\protect\citeauthoryear{Wang and Han}{Wang and Han}{2014}]%
        {wang2014online}
\bibfield{author}{\bibinfo{person}{Xinying Wang} {and} \bibinfo{person}{Min
  Han}.} \bibinfo{year}{2014}\natexlab{}.
\newblock \showarticletitle{Online sequential extreme learning machine with
  kernels for nonstationary time series prediction}.
\newblock \bibinfo{journal}{\emph{Neurocomputing}}  \bibinfo{volume}{145}
  (\bibinfo{year}{2014}), \bibinfo{pages}{90--97}.
\newblock


\bibitem[\protect\citeauthoryear{Wu, Xie, Xinpin, and Song}{Wu
  et~al\mbox{.}}{2012}]%
        {wu2012online}
\bibfield{author}{\bibinfo{person}{Tianshu Wu}, \bibinfo{person}{Kunqing Xie},
  \bibinfo{person}{Dong Xinpin}, {and} \bibinfo{person}{Guojie Song}.}
  \bibinfo{year}{2012}\natexlab{}.
\newblock \showarticletitle{A online boosting approach for traffic flow
  forecasting under abnormal conditions}. In \bibinfo{booktitle}{\emph{2012 9th
  International Conference on Fuzzy Systems and Knowledge Discovery}}. IEEE,
  \bibinfo{pages}{2555--2559}.
\newblock


\bibitem[\protect\citeauthoryear{Yang, Pan, and Tao}{Yang
  et~al\mbox{.}}{2017}]%
        {yang2017robust}
\bibfield{author}{\bibinfo{person}{Haimin Yang}, \bibinfo{person}{Zhisong Pan},
  {and} \bibinfo{person}{Qing Tao}.} \bibinfo{year}{2017}\natexlab{}.
\newblock \showarticletitle{Robust and adaptive online time series prediction
  with long short-term memory}.
\newblock \bibinfo{journal}{\emph{Computational intelligence and neuroscience}}
   \bibinfo{volume}{2017} (\bibinfo{year}{2017}).
\newblock


\bibitem[\protect\citeauthoryear{Zhou, Wu, Zhang, and Zhou}{Zhou
  et~al\mbox{.}}{2016}]%
        {zhou2016minimal}
\bibfield{author}{\bibinfo{person}{Guo-Bing Zhou}, \bibinfo{person}{Jianxin
  Wu}, \bibinfo{person}{Chen-Lin Zhang}, {and} \bibinfo{person}{Zhi-Hua Zhou}.}
  \bibinfo{year}{2016}\natexlab{}.
\newblock \showarticletitle{Minimal gated unit for recurrent neural networks}.
\newblock \bibinfo{journal}{\emph{International Journal of Automation and
  Computing}} \bibinfo{volume}{13}, \bibinfo{number}{3} (\bibinfo{year}{2016}),
  \bibinfo{pages}{226--234}.
\newblock


\bibitem[\protect\citeauthoryear{Zinouri, Taaffe, and Neyens}{Zinouri
  et~al\mbox{.}}{2018}]%
        {zinouri2018modelling}
\bibfield{author}{\bibinfo{person}{Nazanin Zinouri}, \bibinfo{person}{Kevin~M
  Taaffe}, {and} \bibinfo{person}{David~M Neyens}.}
  \bibinfo{year}{2018}\natexlab{}.
\newblock \showarticletitle{Modelling and forecasting daily surgical case
  volume using time series analysis}.
\newblock \bibinfo{journal}{\emph{Health Systems}} \bibinfo{volume}{7},
  \bibinfo{number}{2} (\bibinfo{year}{2018}), \bibinfo{pages}{111--119}.
\newblock


\end{thebibliography}
